\documentclass{article}
\usepackage{spconf,amsmath,graphicx}
\usepackage{multirow}
\usepackage{hyperref}
\usepackage{amssymb}
\usepackage{enumitem}
\usepackage{booktabs}
\usepackage[table,xcdraw]{xcolor}

\definecolor{graycolor}{gray}{0.9}

\title{Identity-Preserving Pose-Guided Character Animation via Facial Landmarks Transformation}
\name{Lianrui Mu\qquad Xingze Zou\qquad Wenjie Zheng\qquad Jiangnan Ye \qquad Haoji Hu}

\address{Zhejiang University}
  
\begin{document}
%
\maketitle
\begin{abstract}
Creating realistic pose-guided image-to-video character animations while preserving facial identity remains challenging, especially in complex and dynamic scenarios such as dancing, where precise identity consistency is crucial. Existing methods frequently encounter difficulties maintaining facial coherence due to misalignments between facial landmarks—extracted from driving videos that provide head pose and expression cues—and the facial geometry of the reference images. To address this limitation, we introduce the \textbf{F}acial \textbf{L}andmarks \textbf{T}ransformation (FLT) method, which leverages a 3D Morphable Model to address this limitation. FLT converts 2D landmarks into a 3D face model, adjusts the 3D face model to align with the reference identity, and then transforms them back into 2D landmarks to guide the image-to-video generation process. This approach ensures accurate alignment with the reference facial geometry, enhancing the consistency between generated videos and reference images. Experimental results demonstrate that FLT effectively preserves facial identity, significantly improving pose-guided character animation models. 
\end{abstract}
\section{Introduction}
\label{sec:intro}

Pose-guided character animation generation has emerged as a pivotal research area, driven by its extensive applications in virtual characters, animation, and video production. By synthesizing videos from reference images, such methods enable the creation of highly personalized content tailored to diverse user requirements.
Early approaches~\cite{yoon2021pose} leveraging Generative Adversarial Networks (GANs), and utilizing UV coordinates to map appearance and labels to a unified representation, achieved notable progress in producing realistic and diverse visual content. However, these methods still exhibit shortcomings in synthesis quality.

\begin{figure}[t]
    \centering
    \includegraphics[width=3.2in]{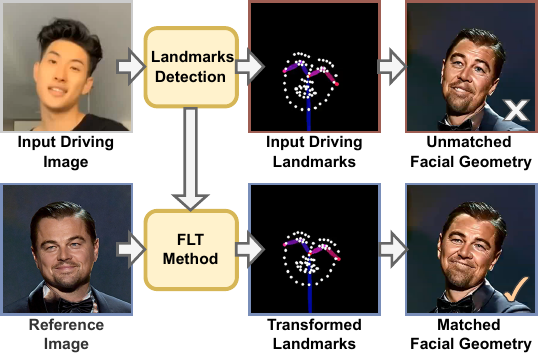}
    \caption{We propose a facial landmark transformation approach using 3D face reconstruction. Our method aligns the driving image's landmarks with a reference face, significantly improving identity consistency in pose-guided generation, even under large facial geometry differences.}
    
    \label{fig_concept}
\end{figure}

Recent diffusion-based video generation techniques~\cite{blattmann2023align} have surpassed typical GAN-based solutions in visual quality. Researchers have leveraged the robust generative capabilities of diffusion models. This has enabled fine-grained control over character movements and scene layouts in pose-guided image-to-video tasks. Notable examples include MagicPose~\cite{chang2023magicpose}, which employs a two-stage training strategy to disentangle appearance and pose, thereby enhancing identity consistency. AnimateAnyone~\cite{hu2024animate} improves temporal coherence through a spatial attention module (ReferenceNet) and a pose guider. ControlNeXt~\cite{peng2024ControlNeXt} introduces a streamlined control architecture, reducing computational overhead and enabling flexible content manipulation. Despite these advancements, identity preservation remains problematic. This issue arises when driving facial landmarks, often extracted from real human videos, deviate significantly from the facial geometry of the reference image. Such mismatches often cause the generated face to inherit features from the driving identity, deviating from the intended reference.

To address this issue, we propose our \textbf{F}acial \textbf{L}andmarks \textbf{T}ransformation (FLT) method, a \emph{training-free}, \emph{plug-and-play} solution that seamlessly integrates into frameworks using facial landmarks as generation conditions. FLT fuses face identity information from a \emph{reference image} with the expressions and poses from a \emph{driving image}. This is achieved by aligning the driving landmarks with the reference identity before feeding them into the video generation model. Leveraging a 3D Morphable Model by Huber \emph{et al.}~\cite{huber2016multiresolution}, FLT parametrizes and adjusts facial shape and expression. It then re-renders the modified 3D face back to 2D, producing transformed landmarks that better capture the reference identity. As illustrated in Fig.~\ref{fig_concept}, FLT excels in scenarios where the driving and reference identities exhibit large facial geometry differences, effectively preserving facial contours and enhancing overall identity retention. We evaluate FLT on two pose-guided animation models, AnimateAnyone~\cite{hu2024animate} and ControlNeXt~\cite{peng2024ControlNeXt}, using the TikTok~\cite{jafarian2021learning} and UBC Fashion~\cite{zablotskaia2019dwnet} datasets. Our experiments confirm that FLT significantly improves identity preservation in challenging motion settings. The main contributions of this work are summarized as follows:

\begin{itemize}[left=0em, labelsep=1em]
\item We propose the \textbf{F}acial \textbf{L}andmarks \textbf{T}ransformation (FLT), providing accurate facial guidance to maintain the consistency of the reference character's identity.
\item We develop a \textbf{training-free} and \textbf{plug-and-play} tool applicable to various pose-guided character animation generation models, effectively enhancing the performance of identity preservation in our experiments. To contribute to the community, we have \textbf{open-sourced our approach}.
\end{itemize}

\section{Related Work}
\label{sec:related_work}

\subsection{3D Morphable Models}

3D Morphable Models (3DMMs)~\cite{10.1145/311535.311556} were originally proposed by Blanz and Vetter to represent and manipulate 3D facial structures by parameterizing both shape and texture. This representation has proven effective for various face-related tasks, such as recognition, expression manipulation, and facial animation, as it allows 3D face reconstruction from single or multiple 2D images.

Subsequent works have combined 3DMMs with deep learning to enhance accuracy and robustness in 3D face reconstruction~\cite{deng2019accurate, booth20173d}. In video generation and editing, 3DMM-based strategies ensure facial coherence across frames by aligning model parameters with target features~\cite{kim2018deep}. This alignment is pivotal in preserving identity, especially when a subject undergoes complex movements or diverse expressions.  

In our work, we build upon a 3DMM framework developed by Huber \emph{et al.}~\cite{huber2016multiresolution}, which incorporates a large-scale 3D scan dataset covering diverse facial geometries. It supports convenient PCA-based shape manipulation and includes an expression blendshape model for linear expression blending. By leveraging these properties, our method transforms 2D landmarks into a 3D face mesh and adapts both shape and expression parameters, ultimately generate transformed landmarks to guide identity-preserving video synthesis.

\subsection{Pose-guided Character Generation}
Recent advancements in pose-guided character generation have made notable progress in addressing facial consistency and identity preservation challenges. Approaches like DreamPose~\cite{karras2023dreampose} incorporate character features and pose embeddings, allowing direct concatenation to guide the overall image structure. DisCo~\cite{wang2024disco} focuses on spliting foreground and background for stable video synthesis but relies solely on pose skeletons, overlooking detailed facial cues.

Recent methods that incorporate 68-point facial landmarks—such as MagicPose~\cite{chang2023magicpose}, AnimateAnyone~\cite{hu2024animate}, and ControlNeXt~\cite{peng2024ControlNeXt} have significantly advanced facial detail retention and identity preservation in pose-guided generation. MagicPose~\cite{chang2023magicpose} adopts a two-stage training strategy that disentangles appearance and pose, thereby enabling high-fidelity animations without the need for fine-tuning. By learning separate representations for identity and motion, MagicPose ensures that changes in pose and facial expressions do not degrade identity consistency. AnimateAnyone~\cite{hu2024animate} introduces a spatial attention module (ReferenceNet) and a pose guider, which work in tandem to maintain detailed appearance features while enforcing accurate pose control across consecutive frames. This design not only improves temporal coherence but also upholds a subject's unique facial traits, thereby strengthening identity preservation throughout the animation sequence. ControlNeXt~\cite{peng2024ControlNeXt} provides a lightweight control mechanism that streamlines the generative architecture. Although primarily focused on computational efficiency and flexible content manipulation, ControlNeXt also enhances facial identity fidelity by injecting pose and motion signals in a way that minimally disrupts the subject's intrinsic facial characteristics.

Although existing methods have made progress, they still face the issue where, when driving landmarks differ from the reference geometry, the generated face often inherits unwanted driving face features. Our Facial Landmarks Transformation (FLT) merges the reference shape with the driving expressions in 3D, thereby boosting identity fidelity in pose-guided video generation.

\begin{figure*}[t]
\centering
\includegraphics[width=7in]{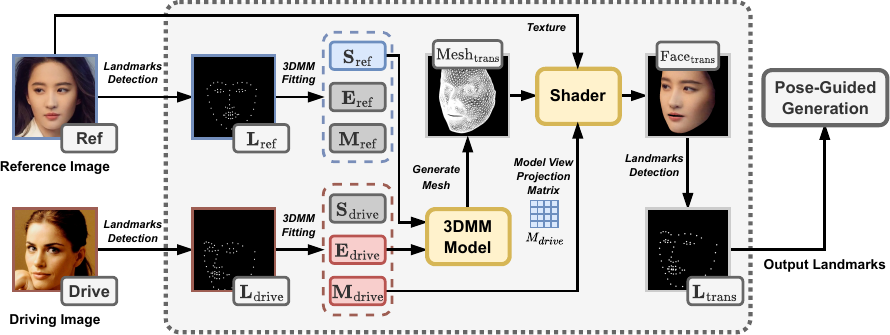}
\caption{\textbf{Overview of the proposed framework.} This pipeline aims to preserve facial identity and consistency in pose-guided video generation. Given a reference image and a driving image, facial landmarks $\mathbf{L}_{\text{ref}}$ and $\mathbf{L}_{\text{drive}}$ are first extracted from both sources. These landmarks are then fitted into a 3D Morphable Model (3DMM) to reconstruct 3D face shapes and capture pose information. Then we use the shape PCA coefficients of the reference image $\mathbf{S}_{\text{ref}}$ and the expression blend shape coefficients of the driving image $\mathbf{E}_{\text{drive}}$ to generate a transformed 3D face mesh $\text{Mesh}_{\text{trans}}$, which is subsequently re-rendered into 2D, ensuring that the reference identity is preserved while adopting the pose and expression dynamics of the driving image. A landmarks detector is applied to extract facial landmarks from the re-rendered face, which are then used as input to guide the video generation model. This approach ensures that the generated video maintains facial consistency and identity throughout dynamic poses and complex motions.}
\label{fig_pipeline}
\end{figure*}

\section{Proposed Method}
\label{sec:method}

Fig.~\ref{fig_pipeline} provides an overview of the proposed pipeline. Our method comprises four main steps: landmark extraction, 3D Morphable Model fitting, identity-preserving reconstruction, and re-rendering with landmark extraction. We then feed the transformed landmarks back into a pose-guided image-to-video generation model to produce identity-consistent animations.

\subsection{Landmark Extraction}
Given two input images---\emph{driving} and \emph{reference}---we first extract their respective 2D facial landmarks, denoted as $\mathbf{L}_{\text{drive}}$ and $\mathbf{L}_{\text{ref}}$, using a standard facial landmark detector~\cite{Dlib}. These landmarks provide the key feature points necessary to perform subsequent 3D reconstructions and alignments in our FLT approach.

\subsection{3D Morphable Model Fitting}
To recover the 3D face geometry from each set of landmarks, we adopt a 3D Morphable Model (3DMM) framework proposed by Huber \emph{et al.}~\cite{huber2016multiresolution} to fit the extracted 2D landmarks, which include a shape PCA model and an expression blend shape model. Specifically, we compute the shape PCA coefficients $\mathbf{S}_{\text{drive}}$ and $\mathbf{S}_{\text{ref}}$, expression blend shape coefficients $\mathbf{E}_{\text{drive}}$ and $\mathbf{E}_{\text{ref}}$, and model-view-projection matrices $\mathbf{M}_{\text{drive}}$ and $\mathbf{M}_{\text{ref}}$.

\subsection{Identity-Preserving Reconstruction}
Once we obtain the 3DMM parameters for both faces, we seek to retain the reference identity while transferring the driving face's expressions. The general 3DMM formulation for a face mesh is:
\begin{equation}
\text{Mesh} = \bar{V} + \sum_{i=1}^{M}{\alpha_i V_{S_i}} + \sum_{j=1}^{K}{\beta_j V_{E_j}},
\end{equation}
where $\bar{V}\in \mathbb{R}^{3N}$ is the mean shape, $N$ is the number of vertices in the mesh. $V_{S_i}$ is the $i$-th shape principal component that describe how the face departs from the mean shape. $V_{E_j}$ is the $j$-th expression blend shape, capturing variations such as smiles, frowns, or mouth opening, representing the dynamic expressions. The $\alpha_i$ and $\beta_j$ are the corresponding shape and expression coefficients, respectively.

To merge the identity of the reference face with the expressions from the driving face, we construct a transformed 3D mesh $\text{Mesh}_{\text{trans}}$, by using only the reference's shape coefficients $\mathbf{S}_{\text{ref}}$  along with only the driving's expression coefficients $\mathbf{E}_{\text{drive}}$. 
\begin{equation}
\text{Mesh}_{\text{trans}} = \bar{V} + \sum_{i=1}^{M}{ \mathbf{S}_{\text{ref}_i}}V_{S_i} + \sum_{j=1}^{K}{ \mathbf{E}_{\text{drive}_j}V_{E_j}}.
\end{equation}

By separating shape from expression, this approach ensures that the resulting 3D mesh $\text{Mesh}_{\text{trans}}$ faithfully reflects the reference person's overall identity while adopting the driving face's facial expression and pose.

\subsection{Rendering and Landmark Extraction}
Directly reading off new 2D landmarks from $\text{Mesh}_{\text{trans}}$ can be unreliable when dealing with complex poses or partial occlusions. Instead, we re-render the face mesh into 2D ensuring consistent viewpoint alignment with the driving image, robustly handling large pose and angle variations. Using the model-view-projection matrix $\mathbf{M}_{\text{drive}}$ estimated from the driving image, we project the $\text{Mesh}_{\text{trans}}$ into 2D face image. During this process, we employ the reference image as a texture source via a simple shader, producing a synthesized 2D face image $\text{Face}_{trans}$ that retains the reference identity while inheriting the expressions and pose from the driving image. 

Finally, we extract updated 2D landmarks from this synthesized face image $\text{Face}_{trans}$, yielding landmark coordinates that precisely encode the target identity's facial structure alongside the driving expressions. These transformed landmarks serve as crucial guiding signals for pose-guided image-to-video generation models, enabling robust facial identity consistency.

\section{Experiment}
\begin{figure*}[t]
\centering
\includegraphics[width=7in]{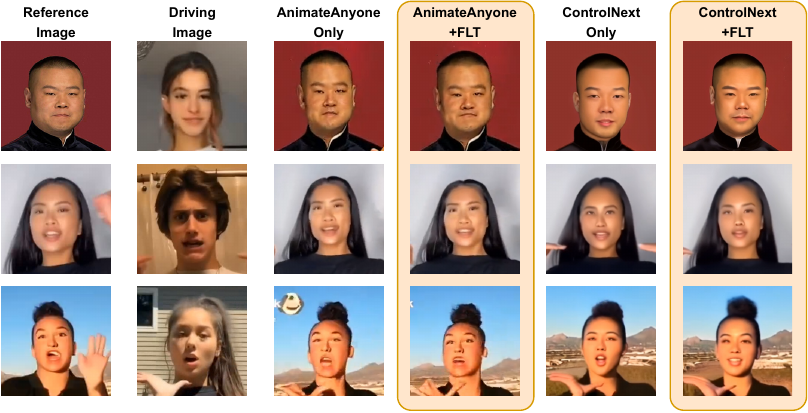}
\caption{We compared the generated images with and without our FLT method when applied to AnimateAnyone~\cite{hu2024animate} and ControlNeXt~\cite{peng2024ControlNeXt}. The results show that our method effectively preserves the reference image's facial features, even with significant differences in facial contours, highlighting FLT's superior identity-preserving performance.}
\label{fig_result}
\end{figure*}

\label{sec:experiment}

\subsection{Experimental Setup}
We evaluate our approach on two publicly available datasets:
TikTok~\cite{jafarian2021learning}, consisting of short challenge videos that often feature dancing, quick head movements, and diverse facial expressions and UBC Fashion~\cite{zablotskaia2019dwnet}, containing high-resolution videos recorded with a mostly static camera, leading to less dynamic facial variations.
From each video, we extract a 1-second clip and use DWPose~\cite{yang2023effective} to obtain the pose skeletons and facial landmarks for every frame in these clips. Additionally, we select one clear facial image from each video to serve as the reference for character animation. To evaluate the effectiveness of our proposed FLT approach, we integrate it into two pose-guided character animation model: AnimateAnyone~\cite{hu2024animate} and ControlNeXt~\cite{peng2024ControlNeXt}. We report results on both TikTok and UBC Fashion to demonstrate the generality of our approach.

\begin{figure}[t]
    \centering
    \includegraphics[width=3.5in]{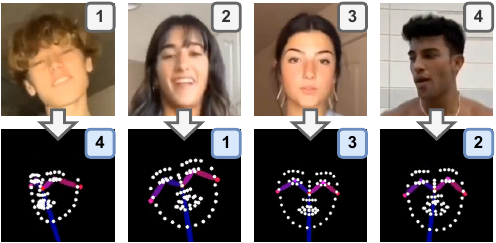}
    \caption{We employ a dataset shuffle procedure when testing FLT performance to evaluate the identity-preserving ability of our method.}
    \label{fig_shuffle}
    \vspace{-0.2cm}
\end{figure}

\subsection{Evaluation Metrics}

We define an optimal benchmark for evaluating our FLT method. Specifically, we consider the matched condition, where the driving landmarks and the reference image are sourced from the same video. This scenario represents the ideal case (upper bound) where facial features perfectly align, providing a reference for the best possible performance of our method. Using the FLT method, we generate videos that align the driving landmarks with the facial features of the reference image. However, standard reconstruction metrics are difficult to apply, as very few datasets contain paired videos of two different subjects with identical poses and facial movements.

To address this limitation, we evaluate identity preservation by comparing the generated frames to the reference image within a dataset shuffle scenario, as illustrated in Fig.~\ref{fig_shuffle}. Let $V_1, V_2, \dots, V_n$ represent $n$ videos from the datasets, and $ \text{ref}_1, \text{ref}_2, \dots, \text{ref}_n$ be one reference image taken from each corresponding video. We define a shuffle function $\sigma$ with a fixed random seed to create a bijection $\sigma: \{V_1, V_2, \dots, V_n\} \to \{\text{ref}_1, \text{ref}_2, \dots, \text{ref}_n\}$ between the set of videos and reference images, ensuring that each $V_i$ is is paired with a potentially mismatched reference image $\text{ref}_{\sigma(i)}$. In this way, we can systematically evaluate how effectively our method preserves the reference identity, even without datasets containing paired videos of two individuals with identical head poses and body movements.

For evaluation, we use two primary metrics: Fréchet Inception Distance (FID)~\cite{heusel2017gans} and facial similarity to test our method under dataset shuffle scenario. For FID, we crop faces from both the reference images and generated frames, then measure the distance between these two distributions to measure how closely the generated faces match the reference. For facial similarity, we employ ArcFace~\cite{deng2019arcface} to extract feature vectors for both the generated and reference faces and use the cosine similarity between these vectors to measure their resemblance. For each generated video, we compute the frame-by-frame cosine similarity with the reference image and take the average over all frames. We also calculate the variance of these similarity scores to evaluate the face consistency.

For a video with $N$ frames, we compute the average similarity $\bar{S}$ and variance $\sigma^2_S$.
\begin{equation}
\bar{S}=\frac{1}{N}\sum_{i=1}^{N}{S_i}\quad \sigma^2_S=\frac{1}{N}\sum_{i=1}^{N}{(S_i-\bar{S})^2}
\end{equation}
Where $S_i$ is the cosine similarity score between the $i$-th frame and the reference face.

\begin{table}[ht]
\setlength{\tabcolsep}{6pt}
\renewcommand{\arraystretch}{1.3}
\begin{tabular}{|c|c|c|c|c|}
\hline
\rowcolor{gray!20} \textbf{Dataset} & \textbf{Methods} & \textbf{Avg $\uparrow$} & \textbf{Var$\downarrow$} & \textbf{FID$\downarrow$}\\
\hline
\multirow{6}{*}{TikTok~\cite{jafarian2021learning}} 
& \textcolor{gray}{CNX Target} & \textcolor{gray}{0.389} & \textcolor{gray}{0.161} & \textcolor{gray}{73.78}\\
\cline{2-5}
& CNX Only & 0.268 & 0.160 & 75.30 \\
\cline{2-5}
& \textbf{CNX+FLT} & \textbf{0.291} & \textbf{0.159} & \textbf{75.29} \\
\cline{2-5}
& \textcolor{gray}{AA Target} & \textcolor{gray}{0.540} & \textcolor{gray}{0.067} & \textcolor{gray}{102.76}\\
\cline{2-5}
& AA Only & 0.384 & 0.060 & 117.99\\
\cline{2-5}
& \textbf{AA+FLT}  & \textbf{0.402} & \textbf{0.060} & \textbf{113.74}\\
\hline
\multirow{6}{*}{\parbox{2cm}{\centering UBC\\Fashion~\cite{zablotskaia2019dwnet}}} 
& \textcolor{gray}{CNX Target} & \textcolor{gray}{0.426} & \textcolor{gray}{0.068} & \textcolor{gray}{46.80}\\
\cline{2-5}
& CNX Only & 0.400 & 0.071 & 48.77\\
\cline{2-5}
& \textbf{CNX+FLT} & \textbf{0.414} & \textbf{0.062} & \textbf{47.28}\\
\cline{2-5}
& \textcolor{gray}{AA Target} & \textcolor{gray}{0.431} & \textcolor{gray}{0.029} & \textcolor{gray}{86.59}\\
\cline{2-5}
& AA Only & 0.414 & 0.029 & 87.87\\
\cline{2-5}
& \textbf{AA+FLT} & \textbf{0.418} & \textbf{0.029} & \textbf{87.49}\\ %
\hline
\end{tabular}
\label{tab:result}
\caption{\textbf{FLT performance on Tiktok and UBC Fashion dataset.} AA stands for AnimateAnyone~\cite{hu2024animate}, and CNX stands for ControlNeXt~\cite{peng2024ControlNeXt}. Avg and Var mean average cosine similarity and the average variance of the similarity on the test dataset. FID stand for Fréchet Inception Distance. \textcolor{gray}{Target} refers to the optimal performance under conditions of complete facial feature matching, serving as the benchmark for comparison.}
\end{table}

\subsection{Performance}
As illustrated in Fig.~\ref{fig_result}, our method successfully preserves facial identity even when the reference and driving images differ significantly in contour and shape. As shown in Tab.~\ref{tab:result}, FLT achieves higher average similarity scores, approaching the target optimal level. It also demonstrates lower variance, signifying stronger identity retention and better temporal consistency. On the TikTok dataset, which features highly dynamic facial expressions, we observe substantial gains. On the less variable UBC Fashion dataset, the improvement is smaller but still notable. Moreover, FID decreases with our method, indicating closer alignment with the reference distribution and overall quality improvement. Altogether, these results confirm that FLT effectively preserves facial identity and produces more coherent animations, outperforming baseline methods in both similarity and consistency.

\section{Conclusion and Limitations}
\label{sec:conclusion}
The proposed Facial Landmarks Transformation (FLT) method integrates landmark extraction, 3D face fitting, identity-expression fusion, and landmark transformation into a unified framework. This enables the generation of transformed landmarks that preserve the reference image's facial identity while adopting the driving face's expressive dynamics. FLT enhances pose-guided character animation generation tasks by delivering precise facial detail guidance, improving identity preservation, and ensuring feature consistency. Moreover, it can be easily integrated into existing generation models to enhance visual coherence.

While FLT offers significant advancements in facial identity consistency, its reliance on accurate landmark detection may face challenges in handling rapid motion or occlusion. In future work, we plan to develop an end-to-end framework to further refine landmark transformations and explore extending FLT to handle full-body skeletons.

\bibliographystyle{IEEEbib}
\bibliography{strings,refs}

\end{document}